\documentclass[11pt]{article}

\usepackage[preprint]{acl}
\usepackage{times}
\usepackage{latexsym}
\usepackage[T1]{fontenc}
\usepackage[utf8]{inputenc}
\usepackage{microtype}
\usepackage{inconsolata}
\usepackage{graphicx}
\usepackage{xcolor}
\usepackage{booktabs}
\usepackage{comment}
\usepackage{amsmath}
\usepackage{xspace}

\newcommand{\cw}[1]{#1}
\newcommand{\FinalNumbers}[1]{#1}
\newcommand{\mh}[1]{#1}
\newcommand{\AM}[1]{#1}

\newcommand{\ITC}{ITC\xspace}

\title{From Retrieval to Weights: Parametric Individualization of Small Language Models with Individual Text Corpora}

\author{
  Christoph Wigbels$^{1,\dagger}$ \quad
  Ali Abusaleh$^{2,\dagger}$ \quad
  Markus T. Jansen$^{1}$ \AND
  Alexander Mehler$^{2}$ \quad
  Markus J. Hofmann$^{1}$ \\[0.5em]
  $^{\dagger}$ \small{Equal\ contribution} \\[0.2em]
 $^{1}$Bergische Universit\"at Wuppertal \quad
 $^{2}$Goethe-Universit\"at Frankfurt \\
  {\small\texttt{\{wigbels, mhofmann, mjansen\}@uni-wuppertal.de}} \\
  {\small\texttt{\{a.abusaleh, mehler\}@em.uni-frankfurt.de}}
}

\begin{document}
\maketitle

\begin{abstract}
We approach a cognitive simulation perspective on episodic and semantic memory in multiple-choice question answering by incorporating text from individual text corpora (\ITC) into retrieval-augmented generation and DoRA fine-tuning.
We web-crawl the search histories of $515$ participants who answered $36$ multiple-choice knowledge items and analyze a stratified subsample of $150$ participants.
For each participant, one DoRA adapter consolidates their \ITC into a small language model (SLM) whose baseline correctness falls below the participants' lowest quartile.
The adapter measurably writes the \ITC into the weights: it fits its own participant's held-out text better than other participants' texts ($d_z{=}1.27$), an individuality effect that increases with \ITC size in rank order.
On the generalized knowledge test, however, the adapter adds knowledge rather than alignment with the individual: log-loss match improves, whereas match accuracy under a bias-corrected PMI readout does not, and retrieval adds nothing on top.
\cw{Our results demonstrate that ITCs can be consolidated into the weights of SLMs, an encouraging basis for individualized tutoring agents, and we discuss how to move from there toward a realistic simulation of episodic and semantic memory at the individual level.}

\end{abstract}

\section{Introduction}
\label{intro}

A key objective in both cognitive science and natural language processing (NLP) is to either reproduce or extract human knowledge from language models (LMs).
Most of this effort operates at the population level:
aggregate human knowledge is recovered from word and text embeddings~\citep{grand2022semantic}, LMs are probed as generic cognitive subjects \citep{binz2023cognitive}, and their outputs are aligned with knowledge distributed over populations \citep{heyueya2024psychometric}.
However, what a person knows varies greatly from one individual to another \citep{rolfhus1999assessing, hambrick2002knowledge}. 
Reproducing knowledge at the individual rather than the aggregate level is therefore a distinct and largely open challenge.

Human knowledge about the world is held in semantic memory, which stores general facts separately from the episodic memory for single personal events
\citep{tulving1972episodic, tulving2002episodic}. 
In psychometric terms, this accumulated knowledge is called crystallized intelligence (Gc). It grows when fluid ability is invested in learning directed toward particular domains, influenced by personality and interests \citep{cattell1963, horn1967age, ackerman1996theory}. 
The Openness-Fluid-Crystallized-Intelligence (OFCI) model frames this as environmental enrichment \citep{ziegler2012openness, trapp2019ofci}. According to the OFCI model, openness leads people to seek out stimulating environments that strengthen fluid ability, which builds crystallized intelligence.  
The internet is seen as one such enriching environment: 
much of what a person comes to know is built by engaging with online content. 
\AM{A person's search history records the content with which they have engaged, thereby tracing the knowledge-building related information they have acquired individually. We use this information to create an individual text corpus (ITC) for each person, as described in \citet{hofmann_individual_2024}. This data structure is central to our work.}

Large language models (LLMs) store substantial factual knowledge \cw{\citep{petroni2019language}}, but this knowledge is generic and not tied to a specific person \cw{\citep{zhang2024personalization}}. 
A prior study investigated whether individual knowledge can be simulated from a person's own data \citep{wigbels2026konvens}. It combined the Google search histories of $316$ participants with retrieval-augmented generation (RAG) to predict each participant's answers to multiple-choice knowledge items.
That study introduced the three answer-alignment benchmarks we reuse here, \textbf{match accuracy}, \textbf{log-loss match}, and \textbf{CK-accuracy} (crystallized-knowledge accuracy; defined in \autoref{subsec:evaluation}); it found a detectable individual signal, with match accuracy above chance, but poor calibration, with low probabilities on the participants' answers.

Retrieval is one route by which a person's knowledge can enter a model. 
Semantic knowledge is not confined to the specific situations or episodes in which it was acquired. Through a process called consolidation, episodic experiences are gradually reorganized into a distributed store of semantic knowledge \cw{\citep{mcclelland1995complementary}}. 
\cw{RAG mirrors the episodic step: it keeps the participant's \ITC in an external store and queries it at inference time.}
The second \AM{step} is to consolidate the participant's \ITC into the \AM{memory} weights themselves. 
We implement this approach using a per-participant DoRA (weight-decomposed low-rank adaptation) adapter \citep{liu2024dora}, a parameter-efficient form of continued pretraining that writes the \ITC into a low-rank weight update while keeping the base model frozen.
This reflects the recent advancement of parameter-efficient fine-tuning as a means of personalizing LLMs for individual users \citep{tan2024democratizing}.
Thus, we reframe the problem of simulating individual knowledge as a problem of parametric individualization. \AM{This approach focuses on whether} a person’s knowledge, once encoded in the model's weights, corresponds to their actual responses.

\AM{Using an \ITC as a proxy for a person's knowledge and mapping it onto the weights of an SLM that has been fine-tuned accordingly} changes what a valid readout must look like. 
Standard forced-choice evaluation reads answers from the first-token probabilities of the option identities \cw{\citep{zheng2024selectors}} \mh{(IDs; i.e., A, B, C, or D, in four-option question answering)}. \AM{This presupposes that the model can follow instructions, which it may not be able to do.}
Continued pretraining on running text involves predicting the next token rather than following instructions. Therefore, for parametric individualization, the choice of readout is constitutive rather than incidental (\autoref{subsec:mcq-related}). 
Therefore, we keep the benchmark definitions of the prior study \citep{wigbels2026konvens}, but we operationalize them over the likelihoods of the options \AM{A--D} in all four design cells (cf.\ \autoref{subsec:evaluation}).
This \AM{shift} from retrieval to \AM{weights} has a corresponding \AM{shift} in measurement, from the answer interface to the answer content.
We test three hypotheses:
\begin{itemize}
\item[\textbf{H1}] \textbf{Parametric individualization:} 
    Adding participant-specific adapters improves the match with participants' answers: across retrieval conditions, the match accuracy increases and the log-loss match decreases compared to the baseline.
\item[\textbf{H2}] \textbf{Adapting versus retrieving:}  The adapter and retrieval reach the individual signal through different routes, so their effects need not be additive. 
    \AM{To test H2, we examine} their interaction: whether writing knowledge into weights substitutes for retrieval (a larger adapter effect without RAG) or complements it (a persistent adapter effect under RAG). 
\item[\textbf{H3}] \textbf{Robustness to \ITC size:}
      The adapter and retrieval effects persist when \ITC size is statistically controlled (the \ITC size varies by several orders of magnitude across participants).
\end{itemize}
    
We evaluate our hypotheses in a $2\times2$ design that crosses the individualized semantic memory model with RAG-based episodic memory retrieval. 
The first factor contrasts the frozen baseline with the baseline carrying the participant's DoRA adapter. 
The second factor contrasts answering without retrieval with the \ITC-based RAG (\ITC-RAG) retrieval of \AM{\citet{wigbels2026konvens}}. 
All four conditions are scored \AM{based on the three answer-alignment benchmarks of \AM{\citet{wigbels2026konvens}}}: match accuracy, log-loss match, and CK-accuracy.

\section{Related Work}
\label{sec:related}

\subsection{Individual Text Corpora and Memory Consolidation}
\label{subsec:IC-related}

The individual text corpus rests on the premise that a large enough sample of
the text a person has engaged with approximates the knowledge that person has
acquired -- an approach introduced for individual reading behavior and scaled to
web-scale search histories by \citet{hofmann_individual_2024}. The study
extended here applied this premise through retrieval, leaving the corpus in an external store \citep{wigbels2026konvens}.

How such an \ITC is queried maps onto the organization of human memory.
Complementary learning systems theory distinguishes a fast episodic system from
a slow neocortical one that consolidates repeated experience into distributed
semantic representations \citep{mcclelland1995complementary}, a distinction
increasingly used to interpret memory in language models, where retrieval acts
as an episodic mechanism supplying context at inference time
\citep{dong2025episodic, fountas2025emllm}.
Consolidating the \ITC into the weights is its semantic counterpart, and the present study compares the two routes with
each other.

\subsection{Parameter-Efficient Fine-Tuning and Personalized LLMs}
\label{subsec:peft-related}

Adapting a language model to an individual user has most often been approached
through the input, by retrieving user-specific documents or conditioning on a
user's history at inference time \citep{zhang2024personalization}. A
complementary approach adapts the parameters instead. Parameter-efficient
fine-tuning (PEFT) makes this tractable at the scale of one model per user,
since only a small adapter is stored per person while the base model is shared
\citep{tan2024democratizing}. Low-rank adaptation (LoRA) is the dominant PEFT family,
and \citet{liu2024dora} extend it with DoRA, which decomposes each weight into
a magnitude and a direction and thereby brings its learning behavior closer to
full fine-tuning. Writing new content into the weights, however, interacts
with what the model already knows. \citet{biderman2024lora} show that low-rank updates acquire less new material
than full fine-tuning but also perturb the base model's existing behavior less,
whereas continued pretraining on a narrow corpus
without replay\cw{, that is, without interleaving general data into the
training stream,} is an established cause of degraded general ability, including
instruction-following \citep{ibrahim2024continual, luo2023forgetting}. This
degradation is a property of the training regime rather than of model size,
with which forgetting does not vary monotonically \citep{ramasesh2022scale,
luo2023forgetting}. We therefore adopt DoRA for its higher adaptation capacity,
appropriate for writing an \ITC into a small base model; the
accompanying loss of instruction-following is not a failure of the model but a
measurement question, which \autoref{subsec:mcq-related} takes up.

\subsection{Measuring Forced-Choice Answer Behavior}
\label{subsec:mcq-related}

How an answer is extracted from an LLM is not neutral. One line of
work shows that forced-choice readouts carry model-side response biases:
LLMs prefer specific option IDs regardless of content, an effect that
survey-style questioning amplifies \citep{dominguezolmedo2023surveys} and
that ablations locate \mh{the decision} in the ID tokens themselves rather than in the
ordering positions of the options \citep{zheng2024selectors}. A second line
separates readout from the answer: \citet{wang2024firsttoken} show that the
option ranked highest by first-token probability frequently differs from the
answer the model generates as text, with the mismatch growing as
instruction-following ability decreases, and text answers prove more
robust to option reordering and question perturbations than first-token
probabilities, including debiased ones \citep{wang2024looktext}. Scoring
option contents by their length-normalized likelihood is the older readout
and remains standard for models that do not follow instructions
\citep{wang2024looktext}. For the present study these findings carry a
design consequence rather than a caveat: adapters obtained by raw continued
pretraining are not instruction-followers by construction, so all four
cells are scored on option contents (\autoref{subsec:evaluation}).

\section{Methods}
\label{sec:methods}

\subsection{Design overview}
\label{subsec:design}

\begin{figure*}[t]
  \centering
  \includegraphics[width=\textwidth]{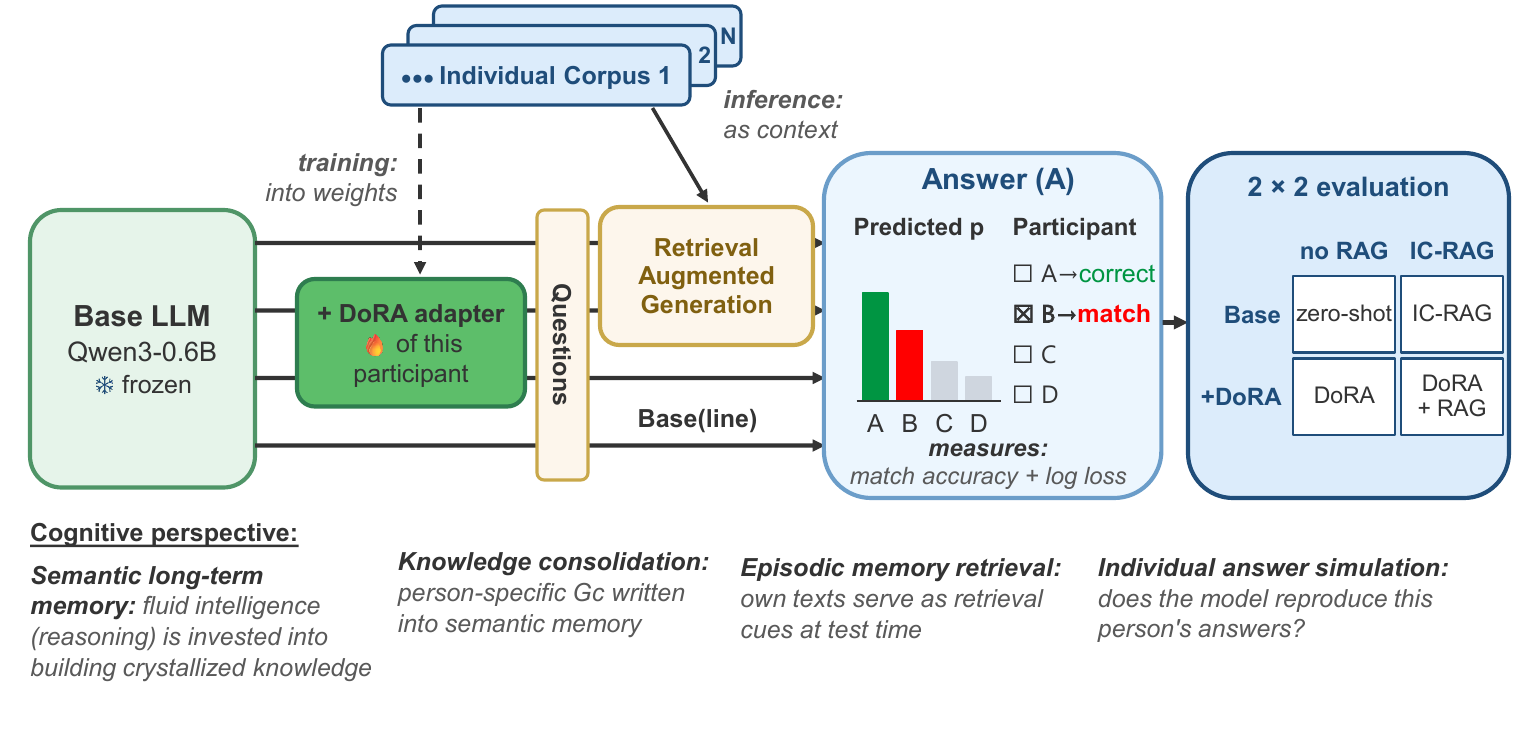}
  \caption{Overview of the study.}
  \label{fig:overview}
\end{figure*}

\autoref{fig:overview} summarizes the study design. We selected \texttt{Qwen3-0.6B} as the base model, since smaller language models have been
found to align better with human knowledge distributions than larger ones \citep{heyueya2024psychometric}. The base model remains frozen in bfloat16; the per-participant adapters are specified in \autoref{subsec:dora-training}.

\subsection{Participants and Knowledge Assessment}
\label{subsec:participants}
Participants were recruited through university flyers, the SONA system, and personal recruitment. Eligibility required fluency in German, active use of a Google account for at least one year, and completion of an online survey of 60 to 90 minutes covering knowledge, personality, interests, fluid intelligence, and demographic information. Participants received course credit or a payment of up to 18~euros. A participant was excluded if they did not provide a valid search history file or if the resulting \ITC fell below the minimum size criterion of 2{,}500 unique word types. Crystallized knowledge was assessed with 36 multiple-choice items, the 12 items of the BEFKI GC-K short scale \citep{schipolowskiBEFKIGCKShort2013} and 24 newly developed items, each with a question stem, four answer options, and exactly one correct answer (example item in Appendix~\ref{app:item}). \mh{After applying these criteria, $515$ participants qualify.} \cw{The present study is based on a stratified subsample of $N{=}150$ of them, drawn across the \ITC size range with outliers trimmed at both ends. Twelve of these participants failed the control questions of the knowledge assessment and are excluded from the answer-level analyses ($N{=}138$); the text-level analyses use all $N{=}150$.}

\subsection{Individual Text Corpora}

\label{subsec:IC-methods}
Each participant's \ITC was assembled from their own Google search history, so that the corpus consists of the running text that participant actually engaged with online. The participant exported this history through Google Takeout, from which we extracted the list of visited URLs. The pages behind these URLs were retrieved with a two-pass scraping pipeline: a first HTTP-extraction pass and a second JavaScript-rendering pass, both routed through rotating proxies, with support for PDF documents. A cleaning filter then removed boilerplate content and discarded non-German text, excluding any sentence with fewer than three German stop words. \cw{English-language content was machine-translated into German with NLLB \citep{nllb2022} beforehand, so that engagement with English pages is retained in German form rather than discarded.} Duplicate documents were removed, so that pages a participant visited repeatedly do not inflate the \ITC.

\subsection{Per-Participant DoRA Adapters}
\label{subsec:dora-training}
Each participant's knowledge is consolidated into the shared base model through a dedicated per-participant DoRA adapter \citep{liu2024dora}, whose higher adaptation capacity at equal rank (\autoref{subsec:peft-related}) suits our setting, where a substantial amount of new content must be written into a small model. With the base weights frozen, the adapter is a compact carrier of what is unique to that person, while the base model holds what is shared across participants: a parametric counterpart to the consolidation of episodic experience into semantic memory \citep{mcclelland1995complementary}. For each participant, one DoRA adapter ($r{=}16$, $\alpha{=}32$, dropout $0.05$, all linear modules) is trained by raw continued pretraining on that participant's cleaned \ITC (learning rate $1.5\times10^{-4}$, constant after $5\%$ warmup, paged 32-bit AdamW, effective batch size of $24$ sequences of $2{,}048$ tokens, bf16), without any synthetic augmentation, using a seeded document-level train/held-out split and early stopping on the held-out loss (epoch ceiling of $10$, patience of $3$). The objective thus writes the knowledge distribution of the \ITC into the weights rather than teaching an answer format; answers are therefore scored over option contents (\autoref{subsec:evaluation}).

\subsection{Evaluation}
\label{subsec:evaluation}
All four cells are scored on the answer-alignment benchmarks of the prior study: match accuracy and log-loss match as primary benchmarks, and CK-accuracy as a secondary one \citep{wigbels2026konvens}. As motivated in \autoref{intro} and  \autoref{subsec:mcq-related}, we keep these benchmark definitions but operationalize them over the option \emph{contents} rather than the option IDs. We therefore score every cell by the length-normalized sequence likelihood of each option's content: the mean per-token log-likelihood of the option text given the question prompt, renormalized over the four options into an answer distribution. 
Match accuracy is then the share of items where the most likely option equals the participant's answer. Log-loss  match is the negative log probability of the participant's option. CK-accuracy is the share of items where the correctness of the model's most likely option matches the correctness of the participant's answer.
\cw{We take as our primary scoring view a domain-conditional pointwise-mutual-information (PMI) variant that additionally subtracts each option's mean per-token log-likelihood under a neutral null prompt (``Antwort:'', German for ``Answer:''), which removes preferences for generic option strings \citep{brown2020language, holtzman2021surface}. The resulting scores are renormalized over the four options in the same way, and the length-normalized (LN) likelihood above then serves as a robustness view}. A permutation-balanced control bank\cw{, a small synthetic item set in which every answer option appears equally often in every position,} confirmed that this readout is content-driven in both model arms (base and adapter) \cw{(control accuracy $0.50$ for the base model and $0.70$--$0.72$ for adapters at a chance level of $0.25$)}, whereas ID-based readouts sit at chance level for the adapters.
\cw{In the retrieval cells, the participant's \ITC is queried with the item stem
through dense retrieval: nomic-embed-text-v1.5 embeddings over the \ITC chunks, with chunking identical to the prior study's retrieval store \citep{wigbels2026konvens}. The five best-matching chunks are inserted into the prompt ahead of the question. All prompts end with the neutral cue ``Antwort:'' and use no chat template, and for a given participant and item the retrieved context is identical in the base and the adapter arm.}
\cw{We split items by participant correctness and decompose the adapter's change in option probability within it. Writing $\delta_x = p_x(\text{adapter}) - p_x(\text{base})$ for the shift on option $x$, \emph{knowledge gain} is, on correctly answered items, $\delta$ on the correct option minus the mean $\delta$ over the three wrong options, the generic injection readout. On incorrectly answered items we report \emph{chosen-option mass}, $\delta$ on the option the participant actually chose, and \emph{error alignment}, that shift minus the mean $\delta$ over the two other wrong options; the latter is positive only when the adapter moves toward this participant's specific error rather than toward plausible distractors in general, isolating person-specific alignment from generic correctness. 
Knowledge gain is thus the graded, likelihood-based analogue of CK-accuracy, whereas error alignment measures what CK-accuracy cannot see, namely which wrong option a participant chooses. All three are computed under both scoring views.}
Since the operationalization differs from that of the prior study, absolute levels are not comparable with the values reported there; all hypothesis tests are within-method contrasts between the four cells, with \ITC size statistically controlled (H3).
\cw{To test whether an adapter carries person-specific rather than generic information, we additionally score it on held-out running text (per-token NLL over packed $2048$-token blocks; protocol in Appendix~\ref{app:nll}).
For each participant we compute the reduction in per-token negative log-likelihood of their own adapter relative to the base model on their own held-out documents (the injection check $g_{\mathrm{own}}$), and on the held-out documents of $20$ seeded partner participants ($g_{\mathrm{foreign}}$).
The individuality contrast $g_{\mathrm{own}} - \bar g_{\mathrm{foreign}}$ is positive only when the adapter fits its own person's unseen text beyond any generic gain; this measurement uses no answer format and provides the foreign reference the answer level lacks.}
\cw{All hypothesis tests are paired within participants: item-level scores are averaged per participant and cell, and the per-participant differences are tested. A Shapiro--Wilk check selects the test, Student's $t$ when normality holds and the Wilcoxon signed-rank test otherwise, and effect sizes are reported as within-subject Cohen's $d_z$, the mean difference divided by the standard deviation of the differences. The H2 interaction is computed per participant as the difference of the adapter effect between the two retrieval conditions. For H3, each per-participant contrast is correlated with $\log_{10}$ \ITC size (Pearson and Spearman), so that an effect carried by \ITC size would appear as a nonzero association.}

\section{Results}

\begin{figure}[t]
  \centering
  \includegraphics[width=\columnwidth]{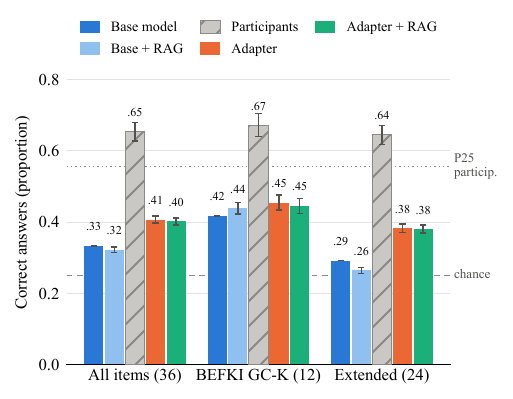}
 \caption{Answer correctness (share of items answered correctly) for the four design cells and the participants, for all $36$ knowledge items and separately for the $12$ public BEFKI GC-K items and the $24$ unpublished extended items ($N{=}138$, PMI scoring view). Error bars are $95\%$ confidence intervals across participants. The dashed line marks the four-option chance level ($0.25$), the dotted line the participants' lower quartile ($0.556$); every model cell falls below it. The base model answers the public items markedly more often correctly than the unpublished ones, whereas the participants do not.}
  \label{fig:befki-correctness}
\end{figure}

\subsection{Parametric Individualization (H1)}
\label{subsec:results-h1}

\begin{figure}[t]
  \centering
  \includegraphics[width=\columnwidth]{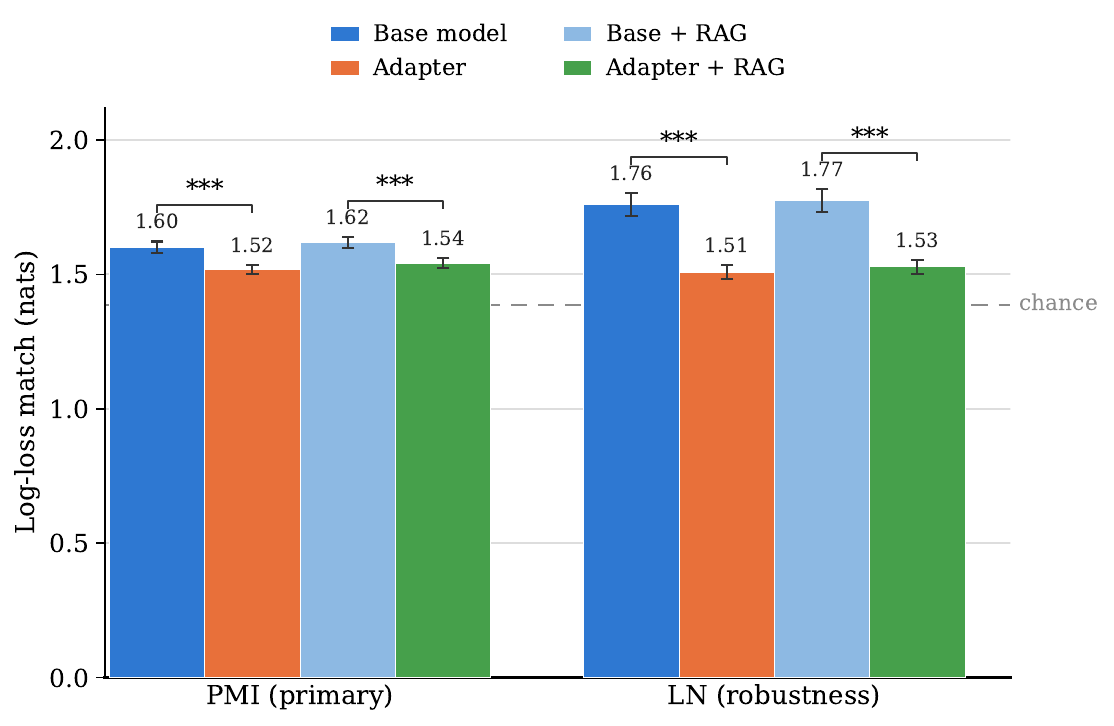}
 \caption{Log-loss match (negative log probability of the participant's own answer; lower is better) for base model and per-participant adapter, without retrieval and with \ITC-RAG, under the primary PMI and the length-normalized (LN) robustness view ($N{=}138$). Bars: cell means with $95\%$ CIs across participants; dashed line: chance level ($-\ln 0.25 \approx 1.39$ nats). Brackets: paired base-vs-adapter contrast per retrieval condition (Shapiro--Wilk-selected $t$/Wilcoxon; $^{***}p<.001$; $d_z = 0.51/0.44$ under PMI, $1.74/1.58$ under LN).}
  \label{fig:logloss-2x2}
\end{figure}

At the text level, the injection check $g_{\mathrm{own}}$ is positive: with the participant's own adapter, the model finds that participant's held-out documents on average $0.380$ nats per token less surprising than the frozen base model does ($d_z = 3.74$, one-sample $t(149) = 45.8$, $p < .001$), so the ITCs have  been written into the weights. The individuality contrast is also positive: an adapter fits its own participant's unseen text by $0.107$ nats per token better than it fits other participants' held-out texts ($d_z = 1.27$; Shapiro--Wilk rejected normality, so we report the Wilcoxon test, $p < .001$). The weights therefore carry an individual signal beyond a generic domain gain.

At the answer level (\autoref{tab:h1-answers}), match accuracy
improves under length normalization ($+0.050$,
$d_z = 0.91$) but is null under PMI ($-0.008$,
n.s.). The length-normalized gain is therefore explained by preferences for
generic option strings, which the PMI view removes, rather than by alignment
with the individual participant. Log-loss match improves in both views, by
$0.248$ nats under LN and $0.079$ nats under
PMI: the individualized model assigns more probability to the answers the
participants actually gave, and \autoref{fig:logloss-2x2} shows this
reduction for each design cell. CK-accuracy rises under LN
($+6.1$ percentage points) but falls under PMI
($-3.3$ points). The correctness split explains this pattern.
On correctly answered items the adapter adds probability to the correct
option (knowledge gain $+0.039$ under PMI), whereas on
incorrectly answered items it removes probability from the option the
participant actually chose (chosen-option mass $-0.043$) and
moves away from that participant's specific error rather than toward it
(error alignment $-0.017$). In sum, the adapter makes the
small model more knowledgeable, and this generic correctness gain carries
the log-loss improvement, while the alignment with the individual person's
answer behavior decreases, in particular with their errors. 
\cw{The item-set split doubles as a contamination control
(\autoref{fig:befki-correctness}): the base model is far better on the $12$ public BEFKI GC-K items than on the $24$ unpublished ones ($.417$ vs.\ $.292$), a gap the participants do not show ($.672$ vs.\ $.645$); the adapter shrinks it to $.072$, so the added knowledge lands disproportionately on items that cannot have been in the base model's pretraining data.}

\begin{table*}[t]
\centering\small
\begin{tabular}{@{}l rrrr rrrr@{}}
\toprule
 & \multicolumn{4}{c}{\textbf{LN}} & \multicolumn{4}{c}{\textbf{PMI}} \\
\cmidrule(lr){2-5}\cmidrule(lr){6-9}
\textbf{Contrast} & \textbf{Base} & \textbf{Adapter} & \textbf{M} & $d_z$ & \textbf{Base} & \textbf{Adapter} & \textbf{M} & $d_z$ \\
\midrule
\textit{All items} \\
\quad Match accuracy   & $0.271$ & $0.321$ & $+0.050^{***}$ & $0.91$  & $0.337$ & $0.329$ & $-0.008$       & $-0.13$ \\
\quad Log-loss match   & $1.767$ & $1.518$ & $+0.248^{***}$ & $1.71$  & $1.610$ & $1.530$ & $+0.079^{***}$ & $0.49$ \\
\quad CK-accuracy & $41.2$ & $47.4$ & $+6.1^{***}$ & $0.95$ & $50.3$ & $47.0$ & $-3.3^{***}$ & $-0.47$ \\
\addlinespace
\textit{Correct items} \\
\quad Knowledge gain   & $0.056$ & $0.082$ & $+0.025^{***}$ & $0.87$  & $0.010$ & $0.049$ & $+0.039^{***}$ & $1.02$ \\
\addlinespace
\textit{Incorrect items\textsuperscript{b}} \\
\quad Chosen-option mass & $0.287$ & $0.267$ & $-0.021^{***}$ & $-0.86$ & $0.274$ & $0.230$ & $-0.043^{***}$ & $-1.14$ \\
\quad Error alignment & $0.054$ & $0.036$ & $-0.018^{***}$ & $-0.53$ & $0.012$ & $-0.004$ & $-0.017^{***}$ & $-0.32$ \\
\bottomrule
\end{tabular}
\caption{H1 answer-level contrasts (adapter vs.\ base), $N{=}138$.
\textit{Note.} Base and Adapter are cell means averaged over the two retrieval
conditions; M is the mean within-participant difference (log-loss match as base
minus adapter, so that positive values mean improvement) and $d_z$ the
within-subject Cohen's $d_z$. All measures are defined in
\autoref{subsec:evaluation}. Chance reference: $0.25$ for match accuracy and
chosen-option mass, $-\ln(0.25) \approx 1.39$ nats for log-loss match, and $0$
for knowledge gain and error alignment; CK-accuracy is in percentage points.
PMI is the primary scoring view, LN (length-normalized) the robustness view.
The paired test is selected per contrast by a Shapiro--Wilk check
($t$/Wilcoxon); $^{*}p<.05$, $^{**}p<.01$, $^{***}p<.001$, unmarked entries are
not significant. Each contrast targets a distinct construct, so $p$-values are
uncorrected. \textsuperscript{b}$N{=}136$: two participants answered all items
correctly.}
\label{tab:h1-answers}
\end{table*}

\subsection{Adapter by Retrieval (H2)}

The adapter-by-retrieval interaction is null in the primary PMI view for both
benchmarks (match accuracy $M = +0.004$, $d_z = 0.06$; log-loss match
$M = -0.008$, $d_z = -0.10$; positive values would mean that retrieval helps
more with the adapter than with the base model). In the LN view, the match
interaction is small and negative ($M = -0.015$, $d_z = -0.23$, $p < .01$;
log-loss match $M = -0.007$, $d_z = -0.09$, n.s.), that is, retrieval adds
slightly less on top of the adapter than on top of the base model.

\label{subsec:results-h2}

\subsection{Robustness to \ITC Size (H3)}
\label{subsec:results-h3}
The person-specific text-level effect shows no linear association with \ITC size ($r = .15$, $p = .069$, $N{=}150$), but a positive rank-order association ($\rho = .29$, $p < .001$): individuality tends to be larger for participants with larger corpora, without scaling linearly in its magnitude. 
The effect itself is not carried by \ITC size, since partialing $\log_{10}$ \ITC size out of the contrast leaves its size essentially unchanged (size-adjusted $d_z = 1.29$ against $1.27$ unadjusted).
The answer-level H1 contrasts are size-independent in the primary PMI view  ($|r| \le .05$), and a single nominal association among the eight answer-level size tests (the match interaction under PMI, $r = .19$, $p = .04$) does not survive any correction for multiple testing. 
The likelihood gain from assigning an \ITC to its own participant is therefore present across the whole sampled \ITC-size range.

\section{Discussion and Outlook}
\cw{The adapters demonstrably write individual information into the weights: each adapter fits its own participant's held-out text better than other participants' texts ($d_z = 1.27$). 
At the answer level, H1 is only partially supported: log-loss match improves in both scoring views, whereas match accuracy does not improve in the primary PMI view, and the correctness split attributes the log-loss gain to generic knowledge rather than to alignment with the individual participant (\autoref{tab:h1-answers}). Retrieval does not interact with the adapter effect (H2), and none of the effects is carried by \ITC size (H3). 
In terms of complementary learning systems, consolidation into the weights thus succeeded for the \ITC itself, while neither route reproduced the participant's individual answer behavior.}

 \mh{Our first reason for choosing the relatively small Qwen3-0.6B base model was a previous study suggesting that smaller models align better with human knowledge distributions \citep{heyueya2024psychometric}. Second, this design choice was motivated by a previous study showing that log-loss match is lowest with small language models~\citep{wigbels2026konvens}. Third, pretrained LLMs hardly forget \cw{their factual knowledge}~\citep{cossu2022continualpretrainingmitigatesforgetting}; as knowledge already present in the weights can hardly be removed again,
\cw{we decided to use a base model that performs below the participant population.
As \autoref{fig:befki-correctness} shows, however, the gap is larger than intended: the base model's correctness ($.33$) falls below the participants' lower quartile ($.56$), and in all conditions the model answered worse than the participants did.}
Therefore, this small language model approach has inherent drawbacks. Although fine-tuning injected individual knowledge, it did not raise individual memory simulation to a level of correctness reached by the participants. We see several potential remedies for this problem and plan to test them in the following order: \\
a) The fine-tuning objective may be informed by log-loss match on a training sample, though this would require more knowledge items for testing. As the BEFKI GC-K items represent the knowledge domains of humanities, as well as social and natural science~\citep{schipolowskiBEFKIGCKShort2013}, one could alternatively think of a  pessimistic leave-one-knowledge-domain-out internal cross-validation, which excludes generic domain gain by the model development strategy, and an external hold-out participant sample for testing. As not all knowledge domains may be represented equally well, such an approach might be framed in a data-driven manner by semantic knowledge clusters and questions designed for these more diverse and divergent knowledge clusters.\\
b) We may choose a larger LLM, but such a model may possess too much knowledge. To remedy this problem, parameters may be noised until an appropriate level of correctness is reached, either for the sample's lowest quartile (combinable with option a), or by selecting an appropriate noise level per participant. The target correctness might be set slightly below the participant's level, such that the \ITC only emphasizes that specific participant's knowledge. Since \autoref{fig:befki-correctness} shows that individual fine-tuning increased \cw{the model's answer correctness} by approximately $7$~percentage points, we may select a level of noise that produces a corresponding baseline underperformance in the base model. \\
c) A psychological theory-driven approach might also suggest using each participant's level of fluid intelligence to select an appropriate level of reasoning for their base model. For instance, Qwen3 provides a wide parameter range from 0.6B to 32B~\citep{qwen3}. Therefore, participants' correctness bins could be mapped against model sizes to select a tailored base model.}

 Though we made substantial progress in terms of log-loss match, our values are not directly comparable with the prior study~\citep{wigbels2026konvens}: it read answers from option-ID probabilities, which carry their own letter biases \citep{zheng2024selectors}, whereas our adapters require scoring over option contents (\autoref{subsec:mcq-related}).
The reliable evidence is within-method: the adapter lowers log-loss match relative to the base model in both scoring views (\autoref{fig:logloss-2x2}), whereas \ITC-RAG retrieval does not.
Our values still remain above the chance-level baseline of $-\ln(0.25) = 1.39$ nats, so calibration to the participant's own answer stays limited.
Because scoring over option contents introduces length and surface-form preferences \citep{brown2020language, holtzman2021surface}, the domain-conditional PMI scoring we adopt removes them, which makes the present evaluation more conservative than a raw content readout.

\cw{At the answer level, the adapter thus dissociates knowledge from person-specific alignment: it adds probability mass to correct options, yet moves away from the specific errors of its participant (\autoref{tab:h1-answers}).
Whether model errors align with human error patterns has so far been examined at the population level \citep{liu2025mistakes}; our result indicates that writing a person's \ITC into the weights does not by itself produce this alignment at the individual level.
Research on transactive memory offers one cautious reading of this dissociation: people offload knowledge to the internet and preferentially remember where to find information rather than the information itself \citep{wegner1987transactive, sparrow2011google, ward2013supernormal}.
An \ITC would then partly record content its participant engaged with but never retained, so the adapter can hold knowledge the person does not.
Whether engagement turns into retained knowledge should further depend on interest, which directs both what people seek out and what they consolidate \citep{hidi2006fourphase, ackerman1996theory, trapp2019ofci}.
This reading remains tentative, since we cannot separate it from a generic route in which some knowledge items benefit from web text as such, independently of individual engagement.
A repeated measurement with new knowledge items generated from each participant's own \ITC could discriminate the two accounts: retained, interest-driven knowledge predicts that participants outperform on items grounded in their own \ITC, particularly where item domain and stated interests match, whereas engagement without retention predicts that they do not.
Such \ITC-derived items would also replace the generalized test with an assessment targeted at each participant's own knowledge. \cw{Such consolidated adapters would also be the natural substrate for individualized tutoring agents: a compact per-user weight store that an agent updates as the user's knowledge develops.}}

To sum up from our complementary learning systems perspective: the prior study individualized through episodic retrieval over a larger LLM, whereas here the individual signal is carried by the semantic route, with \ITC-RAG adding nearly nothing on top.
What the smaller model does not reach is the participants' level of correctness, which the adapter raises but cannot level up.

\section*{Limitations}

\cw{Notwithstanding the contribution of these findings, several limitations must be critically noted. 
First, the present analyses rest on a stratified subsample of $150$ participants that was drawn across the \ITC size range with outliers trimmed at both ends, so that participants with extremely small or extremely large corpora are not represented and every \ITC-size conclusion holds only within this restricted range. As in the prior study, the underlying sample was predominantly young, female, and highly educated, which may limit the generalizability of the results to demographic groups whose search behavior and knowledge profiles differ systematically \citep{wigbels2026konvens}.}

\cw{Second, the study is monolingual by design: all knowledge items were administered in German, and the individual corpora consist of German running text, into which English-language content enters only in machine-translated form. Translation quality therefore shapes part of the training signal, and it remains open whether parametric individualization generalizes to other languages or to users modeled in the language of their original engagement.}

\cw{Third, crystallized knowledge was assessed with a generalized test, the BEFKI GC-K plus the 24 items developed in its format, which ranks individuals on a shared item set \citep{schipolowskiBEFKIGCKShort2013}.
A generalized knowledge test captures only knowledge participants hold in common, so the answer-level individualization results are a lower bound on what an assessment targeted at each participant's own knowledge could show.}

\section*{Ethical considerations and data availability}
We obtained approval from the ethics committee of the Bergische Universit\"at Wuppertal and also thoroughly documented the technical and organizational data protection countermeasures for ITCs as potentially person-related data. Participants agreed to share their Google search histories, also for potential scientific re-use. Though we did everything to anonymize ITCs, e.g.,\ by removing raw URLs or keeping only time stamps relative to assessment times rather than precise surfing times, we cannot fully exclude person identifiability. As our search-based ITCs contain only a subset of web-tracking-based ITCs, person identifiability should be lower, however \citep{deusserBrowsingUnicityLimits2020}. We plan to share these data for scientific re-use through the GESIS data archive, well-protected by a non-disclosure agreement, which excludes re-identification and obligates signers to protect these data. A subset of our participants agreed that their names can be stored with the purpose of testing for person identifiability. We plan a shared task under strict data protection conditions, which allows registered teams to participate. Absence of evidence for person identifiability, however, will not imply evidence for its absence, thus the data will remain protected anyway. Nevertheless, positive evidence would demonstrate the need to protect such data or help to inform more successful anonymization. If you are interested in participating, please contact the authors.


\bibliography{References}

\appendix

\section{Knowledge Item Example} \label{app:item}
The 12 BEFKI GC-K core items follow a four-option format and are documented in \citet{schipolowskiBEFKIGCKShort2013}, whereas the 24 extended items are deliberately not published, to keep them out of public text collections as a contamination control, in line with standard practice for psychological test material. The following core item, reproduced from the public BEFKI GC-K documentation in the GESIS instrument archive ZIS\footnote{\url{https://doi.org/10.6102/zis220}}, illustrates the format; as no official English translation exists, the bracketed translation is our own.

\begin{samepage}
\begin{quote}
Wozu dient die Mitose? \\
{[}What is the function of mitosis?{]} \\[0.5em]
$\Box$~Stoffwechselregulation {[}metabolic regulation{]} \\
$\Box$~Fortpflanzung {[}reproduction{]} \\
$\Box$~Bildung von Keimzellen {[}formation of germ cells{]} \\
$\Box$~\textbf{Zellvermehrung bei Wachstumsvorgängen} {[}cell proliferation during growth{]}
\end{quote}
\end{samepage}

\section{Text-Level Scoring Protocol}
\label{app:nll}
\cw{Held-out running text is tokenized without special tokens, concatenated with an EOS delimiter between documents, and packed into $2048$-token blocks scored independently; a shorter final tail block ($\geq 2$ tokens) is retained.
NLL is summed in fp32 across all blocks, and the first token of every block is never predicted (standard causal shift), so per-token NLL is comparable across participants regardless of corpus length.}

\end{document}